\documentclass[letterpaper, 10 pt, conference]{ieeeconf}

\IEEEoverridecommandlockouts   %
\usepackage[T1]{fontenc}       %

\usepackage{graphicx}
\usepackage{booktabs}
\usepackage{multirow}
\usepackage{colortbl}
\usepackage{xcolor}
\let\labelindent\relax        %
\usepackage{enumitem}
\usepackage{float}
\usepackage{amsmath}
\usepackage{amssymb}
\usepackage{textcomp}     %
\usepackage{caption}      %
\usepackage{cite}         %
\usepackage{eso-pic}

\usepackage[colorlinks,breaklinks,citecolor=blue,linkcolor=blue]{hyperref}

\begin{document}

\title{\LARGE \bf HeadIO: Body-Frame Heading Decomposition for Neural Inertial
Odometry on Head-Mounted Devices}

\author{Taeyoung Yeon$^{1}$, Yiquan Li$^{1}$, Chenfeng Gao$^{1}$, Vasco Xu$^{2}$, and Karan Ahuja$^{1}$%
\thanks{$^{1}$Northwestern University, Evanston, IL, USA. $^{2}$University of Chicago, Chicago, IL, USA.}%
}

\AddToShipoutPictureFG*{\AtPageUpperLeft{\put(0,-28){\makebox[\paperwidth][c]{%
  \parbox{0.8\paperwidth}{\centering\footnotesize This work has been submitted to the IEEE for possible publication.
  Copyright may be transferred without notice, after which this version may no longer be accessible.}}}}}
\maketitle
\thispagestyle{empty}
\pagestyle{empty}

\begin{abstract}
Inertial odometry (IO) using only Inertial Measurement Units (IMUs) enables camera-free head tracking in head-mounted devices such as VR headsets, AR glasses, and earbuds.
Prior IO methods operate in a gravity-aligned world frame, 
which is not yaw-equivariant. Yaw augmentation improves robustness but does not guarantee equivariance; learned equivariant layers achieve it, but EqNIO's full pipeline costs nearly 3$\times$ the FLOPs of ours.
While raw body-frame IMU inputs are inherently yaw-invariant, head tilt redistributes both gravity and motion signals across device axes, making the representation tilt-variant.
We propose \textbf{HeadIO}, which combines a body-frame heading decomposition with a bidirectional Mamba backbone for head-mounted inertial odometry.
Our decomposition achieves tilt invariance by separating gravity-aligned and horizontal components using only the tilt component of device orientation.
We evaluate on four head-mounted datasets and find that HeadIO improves RTE-5s by 8.1\% (0.260\,m $\to$ 0.239\,m) over world-frame methods with learned equivariant layers on Nymeria, the largest benchmark dataset, at nearly 3$\times$ lower FLOPs and 8$\times$ fewer parameters.
HeadIO achieves state-of-the-art performance on our primary metric, RTE-5s, on all Aria-glasses datasets, which natively provide body-frame semantics for our decomposition method. We further evaluate the IMU-only deployment scenario on Nymeria, where HeadIO remains state-of-the-art.
We show that our method relies mainly on tilt accuracy, which can be obtained from the gravity signal; for yaw, a single-coefficient gyroscope correction removes 11.6\% of the yaw error and improves RTE-5s by 2.5\%, and sparse external yaw corrections (for example, one every 30\,s) can improve RTE-5s by 12.0\%.
Our decomposition is lightweight, composes with any backbone, and is suitable for real-time deployment on head-mounted devices.
\end{abstract}

\section{Introduction}
\label{sec:intro}

Inertial Measurement Units (IMUs) are compact, low-power sensors widely used for motion tracking in augmented reality, robotics, and wearable devices. Estimating position from IMU signals requires integrating noisy acceleration and angular velocity, which leads to cumulative drift over time. Visual-inertial methods~\cite{mourikis2007msckf, leutenegger2015okvis, qin2018vinsmono, teed2023deeppatchvisualodometry} mitigate this drift by fusing cameras with IMUs, but cameras introduce power, privacy, and environmental constraints, and fast head motion blurs frames enough to defeat standard visual pose estimation~\cite{chen2025imageimu}, making IMU-only odometry attractive for lightweight head-mounted devices such as AR/VR headsets, smart glasses, and earbuds.
Stable head tracking also underpins egocentric demonstration collection for robot learning~\cite{chi2024umi, kareer2024egomimic}, where any tracking drift propagates into the demonstrated manipulation trajectories.

Learning-based inertial odometry (IO) methods replace hand-tuned heuristics with data-driven motion priors~\cite{yan2018ridi, herath2020ronin, liu2020tlio, jayanth2024eqnio, zhao2025tartan}. These methods predominantly adopt a gravity-aligned world-frame representation, where raw IMU measurements are rotated into a stabilized coordinate frame before being fed to the network. This convention was well-motivated for handheld phones, where non-rigid mounting and varied carrying positions make body-frame signals discontinuous across usage contexts. However, for rigidly head-mounted IMUs, world-frame projection discards the implicit gravity signal $\mathbf{R}_t^\top \mathbf{g}$, an encoding of device attitude. AirIO~\cite{qiu2025airio} showed that preserving this signal in the body frame improves ATE by 66.7\% on drone datasets.

Head-mounted devices present a distinct challenge that neither world-frame nor raw body-frame approaches fully address. Like drones, they have rigidly mounted IMUs with rich 3D dynamics, so body-frame representation is beneficial. Unlike drones, whose tilt is mechanically coupled to locomotion heading, humans can move in any direction regardless of where they are looking, making heading a genuinely unobservable nuisance variable. Prior world-frame methods mitigate heading dependence via yaw augmentation~\cite{liu2020tlio} or learned equivariant layers~\cite{jayanth2024eqnio}, but discard body-frame dynamics. Raw body-frame approaches preserve those dynamics but leave gravity entangled across device axes, producing different feature patterns for the same physical action at different head tilts.

\begin{figure*}[t]
    \centering
    \includegraphics[width=0.80\linewidth]{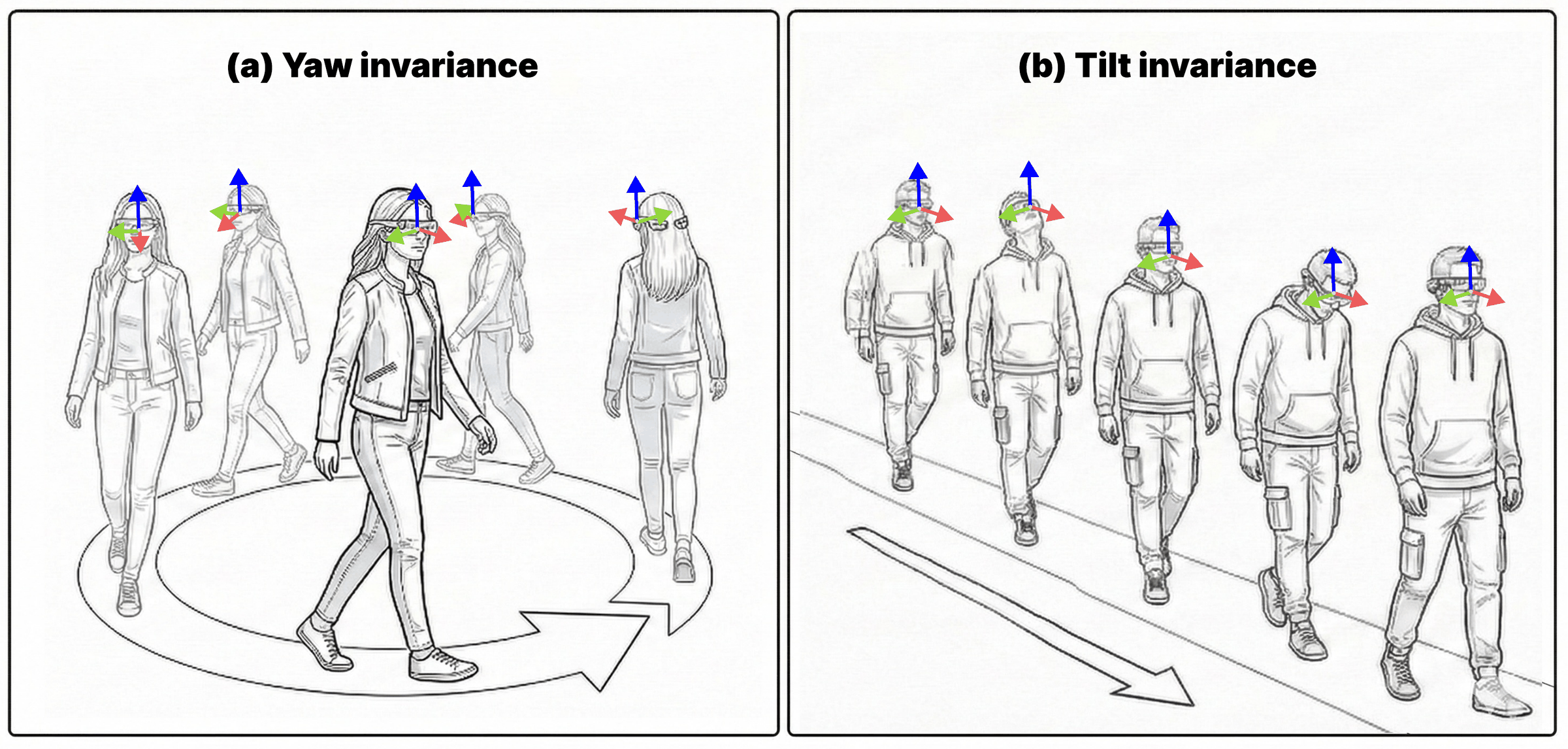}
    \caption{\textbf{Invariance properties of the body-frame heading decomposition.}
    (a)~Body-frame IMU inputs are heading-invariant: the same walking motion produces identical IMU readings regardless of which direction the user faces.
    (b)~Our heading decomposition additionally achieves tilt invariance: as the head pitches or rolls, the gravity-aligned basis adapts so that the forward, lateral, and vertical components remain consistent across head orientations.}
    \label{fig:main}
\end{figure*}

We propose \textbf{HeadIO}, which combines a body-frame heading decomposition with a bidirectional Mamba backbone for head-mounted inertial odometry. Our decomposition is a lightweight analytical preprocessing step that separates body-frame IMU signals into gravity-aligned and horizontal components using only the tilt component of device orientation (Fig.~\ref{fig:main}). This achieves tilt invariance and heading equivariance without learned equivariant layers or architectural constraints, and composes with any backbone. We pair this decomposition with a bidirectional Mamba backbone that uses 7--8$\times$ fewer parameters than TLIO and EqNIO at the shared 3s window.

Through a systematic ablation across four input representations and two backbone architectures, we find that input representation critically interacts with backbone choice.
A Mamba backbone on the standard world-frame input performs 26\% worse than a ResNet (4.09\,m $\to$ 5.16\,m ATE), but with our decomposition it achieves state-of-the-art performance while using no learned preprocessing. This suggests that the sequence model's capacity amplifies representation quality rather than compensating for its deficiencies.

We also evaluate the realistic deployment scenario in which orientation comes only from the IMU, estimated by classical filters. We find that degradation tracks the estimator's tilt error, which the gravity measurement can bound. The four IMU-only estimators have nearly identical yaw error (4.1--4.2$^\circ$) but tilt errors ranging from 0.35$^\circ$ to 5.52$^\circ$. RTE-5s grows with tilt error alone, and HeadIO degrades least, from 0.275\,m to 0.436\,m (1.6$\times$), compared to 2.2$\times$ for TLIO and 3.3$\times$ for EqNIO.

Our key contributions are:
\begin{enumerate}[nosep, leftmargin=*]
    \item We propose a body-frame heading decomposition that disentangles gravity, forward, and lateral components via analytical preprocessing, without learned equivariant architectures.
    \item We conduct ablations across four input representations and two backbone architectures, showing that our preprocessing benefits both the ResNet and the bidirectional Mamba backbone.
    \item HeadIO achieves state-of-the-art RTE-5s, our primary metric, on all three Aria-glasses datasets of the four we evaluate, at nearly 3$\times$ lower FLOPs and 8$\times$ fewer parameters than world-frame learned equivariant methods.
    \item We conduct an orientation-accuracy study under IMU-only deployment, showing that accuracy relies mainly on tilt, which the gravity measurement provides, while yaw is handled by a single-coefficient gyroscope correction (improving RTE-5s by 2.5\%) and sparse external yaw fixes (improving RTE-5s by 12.0\% when given every 30\,s).
\end{enumerate}

\section{Related Work}
\label{sec:related}

\subsection{Inertial Odometry for Human Motion}

Classical methods that use IMUs for human navigation are typically based on strapdown inertial navigation systems (SINS). Traditional SINS methods~\cite{savage1998strapdown_attitude,titterton2004strapdown} estimate motion through double integration of inertial measurements. However, the accumulated integration error leads to rapid drift in practice. To mitigate this issue, Pedestrian Dead Reckoning (PDR) methods introduce
domain-specific constraints based on human walking patterns.
Representative techniques include Zero-Velocity Updates (ZUPT)~\cite{foxlin2005pedestrian},
Zero Angular Rate Updates (ZARU)~\cite{rajagopal2008pdr}, and heuristic drift reduction (HDR)~\cite{borenstein2009heuristic}.

These constraints assume the sensor moves with the walking body. ZUPT requires the stance-phase stillness of a foot-mounted IMU~\cite{foxlin2005pedestrian}, and HDR assumes the heading stays constant during straight-line walking~\cite{borenstein2009heuristic}. A head-mounted IMU on consumer devices such as AR glasses satisfies neither assumption. Unlike the foot, the head does not become stationary at each step, and it can turn freely while the body walks straight or stands still.

More recently, neural inertial odometry approaches have been proposed to replace
hand-crafted heuristics with learned motion priors, directly regressing displacement
or velocity from sequences of IMU measurements. RIDI~\cite{yan2018ridi} first identified that body-frame IMU signals become discontinuous when a handheld phone changes orientation, and proposed regressing velocity in a gravity-stabilized frame to correct low-frequency drift. RoNIN~\cite{herath2020ronin} formalized this with a Heading-Agnostic Coordinate Frame (HACF), gravity-aligning IMU measurements and augmenting with random yaw rotations. TLIO~\cite{liu2020tlio} extended this convention to head-mounted devices, predicting displacement in a global gravity-aligned frame. RIO~\cite{zhou2021riorotationequivariancesupervisedlearning} proposed a self-supervision objective that enforces rotation equivariance in the network to alleviate yaw unobservability. EqNIO~\cite{jayanth2024eqnio} addressed the residual yaw ambiguity via learned Vector Neuron layers that achieve $O(2)$ equivariant canonicalization, but retained gravity-alignment as a foundational assumption.

Most of the above work assumes a gravity-aligned world frame. For rigidly head-mounted IMUs, however, world-frame projection discards the implicit gravity signal $\mathbf{R}_t^\top \mathbf{g}$ and renders displacement targets heading-dependent. None of these methods re-evaluate whether world-frame representation is optimal when the IMU is rigidly mounted.

\subsection{Coordinate Frames for Rigid-Body IO}

For platforms with rigidly mounted IMUs, the choice of coordinate frame changes what the network can learn. DIDO~\cite{zhang2022dido} models quadrotor dynamics in the body frame, where tilt-velocity coupling implicitly resolves heading. Cioffi~et~al.~\cite{cioffi2023learned} show that collective thrust measurements provide a sufficient body-frame prior for drone IO, replacing the need for explicit heading estimation. DIVE~\cite{bajwa2024dive} regresses 3D velocity from IMU history alone for quadrotors, showing that body-frame kinematics provide a sufficient motion prior without additional sensors. AirIO~\cite{qiu2025airio} demonstrated that body-frame representation preserves attitude observability for rigidly mounted UAV IMUs. The term $\mathbf{R}_t^\top \mathbf{g}$ couples linearly with device attitude, a learning signal that world-frame projection destroys, and preserving it improved average ATE by 66.7\% across three drone datasets.

However, these body-frame methods benefit from a property specific to drones: tilt direction is mechanically coupled to locomotion heading, so heading is implicitly resolved. Humans can move in any direction regardless of head orientation. A person may walk forward, sidestep, or turn while looking in an arbitrary direction. This decoupling makes heading a genuinely unobservable degree of freedom for head-mounted IMUs that raw body-frame representations cannot resolve. Our decomposition factors out heading analytically and removes tilt variation from the input, without learned equivariant layers or platform-specific dynamic priors.

\subsection{Backbone Architectures}

ResNet~\cite{he2016resnet} has been the dominant backbone for neural IO since RoNIN~\cite{herath2020ronin} established it as best among ResNet~\cite{he2016resnet}, LSTM~\cite{hochreiter1997lstm}, and TCN~\cite{bai2018tcn}. TLIO~\cite{liu2020tlio} and EqNIO~\cite{jayanth2024eqnio} both inherit this choice. Transformer-based backbones ~\cite{vaswani2017transformer} have since emerged: CTIN~\cite{rao2022ctin} augmented a ResNet encoder with local and global self-attention, and iMoT~\cite{nguyen2025imot} proposed a full encoder-decoder Transformer with cross-modal attention between accelerometer and gyroscope. %
All prior work assumes a fixed input representation, and none examines how representation choice interacts with backbone capacity.

Although Mamba~\cite{gu2023mamba} is a newer approach for temporal modeling, we found it performs worse than a ResNet on the standard world-frame input. With our decomposition, the same backbone outperforms prior methods.

\subsection{Orientation Error in IO}

Learning-based IO methods that take device orientation as an input depend on the quality of IMU-only orientation estimation at deployment. Brossard et al.~\cite{brossard2020denoising} trained a network that removes noise and bias from raw gyroscope measurements, reducing orientation drift. Classical complementary filters such as Mahony~\cite{mahony2008nonlinear} bound tilt error using the accelerometer, but yaw remains unobservable without an external reference such as a magnetometer~\cite{li2026mario} or a visual compass~\cite{kosecka2002video}. None of these works examine how orientation error propagates into learned IO. We measure this propagation across four IMU-only estimators, trace the filter's systematic yaw error to a gyroscope scale-factor term, and show that a yaw fix every 30\,s almost matches the accuracy obtained with ground-truth orientation.

\section{Methods}
\label{sec:methods}

\begin{figure*}[t]
    \centering
    \includegraphics[width=0.80\linewidth]{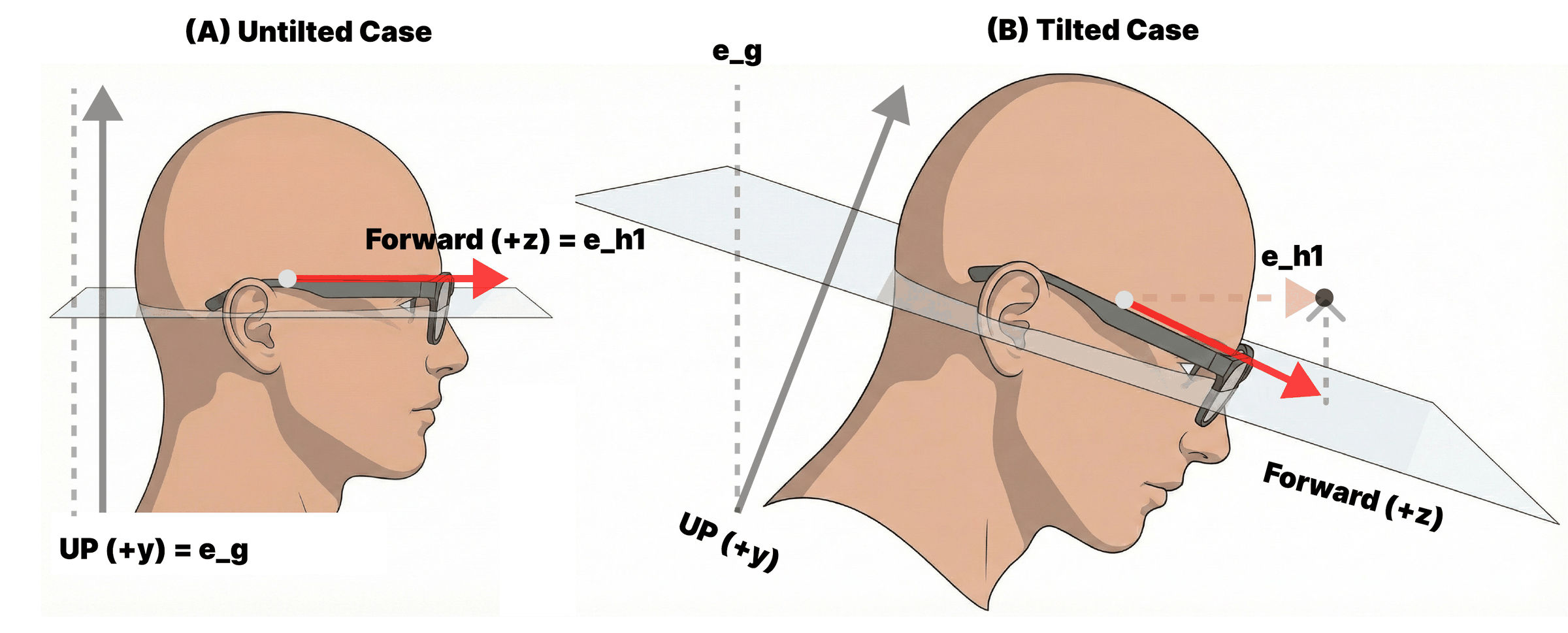}
    \caption{Body-frame heading decomposition. (A) When the head is upright, the body-frame up axis (+y) aligns with gravity direction $\mathbf{e}_g$, and the forward axis $\mathbf{f}_b$ coincides with $\mathbf{e}_{h1}$. (B) When the head tilts, $\mathbf{f}_b$ gains a gravity-aligned component, causing gravity leakage into the forward channel. Our decomposition projects $\mathbf{f}_b$ onto the plane perpendicular to $\mathbf{e}_g$ (Eq.~\ref{eq:basis}), recovering a tilt-invariant forward direction $\mathbf{e}_{h1}$.}
    \label{fig:decomp}
\end{figure*}

\subsection{Body-Frame Heading Decomposition}
\label{sec:decomposition}

An IMU rigidly mounted on a head-mounted device measures acceleration and angular velocity in the device body frame $\mathcal{B}$:
\begin{align}
    \mathbf{a}^{\mathcal{B}}_t &= \mathbf{R}_t^\top \dot{\mathbf{v}}^{\mathcal{W}}_t
      + \mathbf{R}_t^\top \mathbf{g}
      + \mathbf{b}_a + \mathbf{n}_a,
      \label{eq:body_acc} \\
    \boldsymbol{\omega}^{\mathcal{B}}_t &=
      \boldsymbol{\omega}^{\mathrm{true}}_t
      + \mathbf{b}_g + \mathbf{n}_g,
      \label{eq:body_gyro}
\end{align}
where $\mathbf{R}_t \in \mathrm{SO}(3)$ is the device orientation (body $\to$ world), $\mathbf{g}$ is the upward gravity reaction in the world frame (an accelerometer at rest measures $+\|\mathbf{g}\|$), $\mathbf{b}_a, \mathbf{b}_g$ are accelerometer and gyroscope biases, and $\mathbf{n}_a, \mathbf{n}_g$ denote sensor noise. We assume $\mathbf{R}_t$ is available from the on-device orientation filter.

Raw body-frame signals preserve the tilt signal $\mathbf{R}_t^\top \mathbf{g}$ that world-frame projection discards, but head tilt redistributes both gravity and motion signals across device axes, making the representation tilt-variant. Our decomposition separates gravity-aligned and horizontal components, then constructs a heading-invariant basis for the horizontal plane.

\paragraph{Gravity--horizontal separation}
The gravity direction in the body frame
$\mathbf{e}_g = \mathbf{R}_t^\top \hat{\mathbf{g}}$
separates each measurement into a gravity-aligned scalar and
a horizontal remainder:
\begin{align}
    a_g &= \mathbf{a}^{\mathcal{B}}_t \cdot \mathbf{e}_g, &
    \mathbf{a}_h &= \mathbf{a}^{\mathcal{B}}_t
      - a_g \, \mathbf{e}_g, \nonumber\\
    \omega_g &= \boldsymbol{\omega}^{\mathcal{B}}_t
      \cdot \mathbf{e}_g, &
    \boldsymbol{\omega}_h &= \boldsymbol{\omega}^{\mathcal{B}}_t
      - \omega_g \, \mathbf{e}_g.
    \label{eq:grav_horiz}
\end{align}
The scalar $a_g$ captures gravity and vertical acceleration
($a_g \approx \|\mathbf{g}\|$ at rest). The horizontal remainders $\mathbf{a}_h, \boldsymbol{\omega}_h \in \mathbb{R}^3$ lie in the plane perpendicular to $\mathbf{e}_g$ but lack a stable basis across head orientations.

\paragraph{Heading-invariant basis construction}
We project the device forward axis $\mathbf{f}_b$, the wearer's gaze direction, onto the plane perpendicular to $\mathbf{e}_g$ (Fig.~\ref{fig:decomp}):
\begin{equation}
    \mathbf{e}_{h1} = \mathrm{normalize}\!\left(\mathbf{f}_b
      - (\mathbf{f}_b \cdot \mathbf{e}_g)\,\mathbf{e}_g\right),
    \qquad
    \mathbf{e}_{h2} = \mathbf{e}_g \times \mathbf{e}_{h1}.
    \label{eq:basis}
\end{equation}
Since $\mathbf{e}_g = \mathbf{R}_\mathrm{tilt}^\top
\mathbf{R}_\psi^\top \hat{\mathbf{g}}
= \mathbf{R}_\mathrm{tilt}^\top \hat{\mathbf{g}}$
(yaw rotates about $\hat{\mathbf{g}}$, leaving it invariant), the basis $\{\mathbf{e}_g, \mathbf{e}_{h1}, \mathbf{e}_{h2}\}$
depends only on tilt, not heading.

\paragraph{Scalar projection}
Projecting horizontal vectors onto this basis yields the
full per-timestep input
$\mathbf{x}_t = [a_g,\;
\mathbf{a}_h \!\cdot\! \mathbf{e}_{h1},\;
\mathbf{a}_h \!\cdot\! \mathbf{e}_{h2},\;
\omega_g,\;
\boldsymbol{\omega}_h \!\cdot\! \mathbf{e}_{h1},\;
\boldsymbol{\omega}_h \!\cdot\! \mathbf{e}_{h2}]
\in \mathbb{R}^6$.

\paragraph{Exact invariances}
Because dot products are rotation-invariant, each feature can be rewritten in world coordinates, e.g.\ $a_g = (\mathbf{R}_t^\top\mathbf{a}^{\mathcal{W}}_t)\cdot(\mathbf{R}_t^\top\hat{\mathbf{g}}) = \mathbf{a}^{\mathcal{W}}_t\cdot\hat{\mathbf{g}}$: the six scalars are the world-frame specific force and angular rate projected onto the gravity axis and the horizontal gaze direction. The features use only body-frame IMU signals and the canonical basis, so the world yaw orientation is never used to transform the input IMU, unlike prior world-frame methods~\cite{liu2020tlio, jayanth2024eqnio}. The same physical motion with the same horizontal gaze heading also produces identical features at any head tilt, making the representation tilt-invariant. Raw body-frame input~\cite{qiu2025airio} does not have this property: when the head tilts, gravity redistributes across the device axes, and a resting accelerometer reading of $(0, 9.81, 0)$ becomes, for example, $(0, 6.9, 6.9)$ at a 45$^\circ$ tilt. EqNIO learns an $O(2)$-canonical horizontal frame with Vector Neuron layers~\cite{jayanth2024eqnio}. We instead define this canonical frame analytically, from gravity and the wearer's gaze direction, without any learned parameters. (A fallback axis handles the degenerate case of near-vertical gaze.)

\subsection{Displacement Prediction and Trajectory Reconstruction}
\label{sec:reconstruction}

The network predicts decomposed displacement
$[\hat{d}_g, \hat{d}_{h1}, \hat{d}_{h2}]$ from a window
of input features. We reconstruct body-frame displacement
$\hat{\mathbf{d}}^{\mathcal{B}} =
\hat{d}_g \, \mathbf{e}_g +
\hat{d}_{h1} \, \mathbf{e}_{h1} +
\hat{d}_{h2} \, \mathbf{e}_{h2}$
using the precomputed basis, then rotate to world frame:
$\hat{\mathbf{d}}^{\mathcal{W}} =
\mathbf{R}_t \, \hat{\mathbf{d}}^{\mathcal{B}}$,
with the basis and orientation taken at the final window sample.
Heading enters only at the final rotation, not during
feature computation or learning.

\subsection{Backbone}
\label{sec:backbone}

HeadIO processes a window of $T{=}600$ IMU samples at 200\,Hz (3\,s) and predicts the displacement over the final 1\,s. The six decomposition scalars are embedded by a strided convolution (kernel 7, stride 2) into 300 tokens of width 128, followed by three pre-norm residual bidirectional Mamba blocks~\cite{gu2023mamba} (state dimension 16, expansion 2); each block sums a forward and a time-reversed pass, giving full-window context at linear cost. The output sequence is average-pooled to eight temporal positions and decoded by two MLP heads into the displacement mean and its diagonal log-standard deviation. The complete network has 0.97M parameters, independent of window length. We fix the window at 3\,s for all methods based on a window ablation (Sec.~\ref{sec:ablations}). At this shared window, the ResNet baselines require more parameters and compute than at their default 1s windows. TLIO grows from 5.4M to 7.0M parameters and EqNIO from 6.0M to 7.6M, so HeadIO is 7.2$\times$ and 7.8$\times$ smaller, respectively.

\subsection{Yaw Error and Sparse Correction}
\label{sec:yaw_model}

We assume a deployment scenario where orientation comes from an IMU-only filter. The gravity measurement from the accelerometer can bound the filter's tilt error, while yaw remains unbounded from the IMU. Gyroscopes have a small scale-factor error: they report each rotation slightly larger or smaller than it is~\cite{titterton2004strapdown}, so the integrated yaw error grows with the accumulated turn angle. We correct this with a single coefficient, fit on the training split and applied unchanged at test time. The remaining yaw error can only be removed by an external reference. We model such a reference as a yaw fix, resetting the filter's heading to the ground-truth yaw every $\tau$ seconds while leaving its tilt untouched (an oracle for any external heading source), and sweep $\tau$ to measure how sparse these fixes can be.

\section{Experiments}
\label{sec:experiments}

\subsection{Datasets}
\label{sec:datasets}

We evaluate on four head-mounted IMU datasets with varying
motion complexity, sensor hardware, and tilt characteristics.

\paragraph{Coordinate Frame Convention}
We define the body frame as Project Aria's Central Pupil Frame (CPF)~\cite{engel2023aria} for consistency across different datasets. CPF is a device-centric coordinate system defined as ($+X$ left, $+Y$ up, $+Z$ forward). The CPF forward axis ($+Z$) serves as $\mathbf{f}_b$ in our heading decomposition, so using CPF as the body frame ensures a fair comparison: all configurations receive the same body-frame IMU signals, differing only in how they are preprocessed. We define the world frame as a $Y$-up gravity-aligned frame, following RoNIN's heading-agnostic convention~\cite{herath2020ronin} with $Y$-up chosen to match CPF, so that body-to-world rotations involve no axis permutation. Nymeria, AEA, and LaMAria natively provide CPF-aligned data.

\paragraph{Nymeria~\cite{ma2024nymeria}}
Nymeria contains 1200 sequences from 264 participants across 50 indoor and
outdoor locations, totaling over 300 hours. Activities range from athletic motion and cycling to cooking and social interaction, producing diverse and dynamic head motion (tilt SD: 14.54\textdegree). IMU and ground-truth trajectories are recorded at 1\,kHz from Project Aria glasses via Aria Machine Perception Services. We use an 80/10/10 train/validation/test split.

\paragraph{Aria Everyday Activities (AEA)~\cite{lv2024aria}}
AEA contains 143 sequences across five indoor locations, totaling 7.3 hours of household activities such as cooking, tidying, and
social interaction. AEA is recorded with the same Aria hardware and ground-truth pipeline as Nymeria (tilt SD: 11.20\textdegree). We evaluate
Nymeria-trained models on the full AEA dataset without fine-tuning to test cross-dataset generalization.

\paragraph{TLIO~\cite{liu2020tlio}}
The TLIO dataset contains approximately 60 hours of pedestrian walking recorded with a Bosch BMI055 IMU on a custom headset (tilt SD: 12.93\textdegree); ground truth comes from visual-inertial odometry, and we follow the train/val/test protocol of prior work~\cite{liu2020tlio, jayanth2024eqnio}. Unlike the Aria datasets, whose CPF convention semantically defines the body axes, TLIO provides IMU measurements in a gravity-aligned world frame together with a world-to-device orientation; the device frame itself carries no documented axis convention relative to the head. We recover body-frame signals through the provided orientation and align them to a CPF-like frame empirically, from the gravity direction and walking-heading statistics.
A residual rotation that is consistent between training and evaluation is absorbed by the network, so the reconstruction itself does not disadvantage any method; what it cannot recover is a true gaze axis, the heading anchor our decomposition uses.

\paragraph{LaMAria~\cite{pataki2025lamaria}}
LaMAria provides long Aria recordings in industrial and city-scale environments with survey-grade control points and official pseudo-ground-truth. We evaluate the 10 egocentric sequences with released pseudo-ground-truth (tilt SD: 27.71\textdegree), applying pre-trained models from Nymeria without fine-tuning as in the AEA evaluation. We exclude the controlled experimental sequences which are recorded with the device mounted on a movable platform rather than worn on the head~\cite{pataki2025lamaria}.

\subsection{Evaluation Metrics}
\label{sec:metrics}

We report four metrics. \textbf{Absolute Trajectory Error (ATE)} is the root-mean-square position error over the trajectory, following TLIO's evaluation~\cite{liu2020tlio}; trajectories share the initial pose, so no alignment is applied. \textbf{Relative Translational Error at $\Delta t$ seconds (RTE-$\Delta t$)} is the root-mean-square displacement error over fully sliding $\Delta t$-second windows (one-sample stride on trajectories resampled to a uniform 200\,Hz), following AirIO's dense relative-error protocol~\cite{qiu2025airio}. The relative pose error in TLIO's evaluation code instead aligns each window by its initial yaw, which would remove the orientation-error effects we study from the metric. \textbf{Drift Rate} is the final position error as a percentage of ground-truth path length. Per-sequence values are aggregated as the mean over test sequences. On LaMAria, ATE follows the dataset's official protocol (Umeyama-aligned, 2D), marked $^\dagger$ in Table~\ref{tab:main_results}; all other datasets use unaligned ATE.
We prioritize RTE-5s as the primary metric, since ATE averages are dominated by long trajectory sequences. For example, the longest 10\% of Nymeria test sequences (averaging 627\,m of path against a median of 260\,m) account for 26\% of total ATE, while RTE-5s evaluates displacement accuracy over fixed-length windows, allowing fair comparison across datasets of different trajectory lengths.

\subsection{Implementation Details}
\label{sec:implementation}

\paragraph{Training}
Training follows the TLIO protocol: mean-squared error on the decomposed displacement for the first 10 epochs, then a diagonal-Gaussian negative log-likelihood that also trains the uncertainty head. We use Adam~\cite{kingma2017adammethodstochasticoptimization} (learning rate $10^{-4}$, gradient clipping 0.1) with ReduceLROnPlateau (factor 0.1, patience 10), batch size 256, and 50 epochs on a single NVIDIA A40 GPU. All methods are trained with the same datasets, augmentation, and evaluation protocol at the shared 3s window.

\paragraph{Computational cost}
HeadIO has 0.97M parameters and processes a 3s window in 444\,MFLOPs. While TLIO requires 7.00M parameters and 220\,MFLOPs, EqNIO adds learned Vector Neuron layers on the same ResNet and reaches 7.59M parameters and 1227\,MFLOPs, so HeadIO uses 7--8$\times$ fewer parameters than both baselines and 2.8$\times$ less compute than EqNIO, at twice the compute of the plain ResNet.

\section{Results}
\label{sec:results}

\subsection{Main Results}

\begin{table}[t]
\centering
\caption{Main results on the four head-mounted datasets (mean over test sequences; best per metric in \textbf{bold}). RTE and ATE in meters, drift in percent.}
\label{tab:main_results}
\footnotesize
\setlength{\tabcolsep}{3.5pt}
\begin{tabular}{ll cccc}
\toprule
\textbf{Dataset} & \textbf{Method}
& \textbf{RTE-5s} $\downarrow$
& \textbf{RTE-1s} $\downarrow$
& \textbf{ATE} $\downarrow$
& \textbf{Drift\%} $\downarrow$ \\
\midrule
\multirow{3}{*}{Nymeria}
  & TLIO   & 0.278 & 0.101 & 4.093 & 2.17 \\
  & EqNIO  & 0.260 & 0.093 & 3.577 & 1.96 \\
  & HeadIO & \textbf{0.239} & \textbf{0.086} & \textbf{3.258} & \textbf{1.79} \\
\midrule
\multirow{3}{*}{AEA}
  & TLIO   & 0.241 & 0.088 & 0.811 & 5.24 \\
  & EqNIO  & 0.220 & 0.081 & \textbf{0.797} & \textbf{5.14} \\
  & HeadIO & \textbf{0.213} & \textbf{0.077} & 0.819 & 5.81 \\
\midrule
\multirow{3}{*}{LaMAria}
  & TLIO   & 0.863 & 0.255 & 5.949$^\dagger$ & 5.55 \\
  & EqNIO  & 0.804 & \textbf{0.248} & 4.129$^\dagger$ & 4.73 \\
  & HeadIO & \textbf{0.761} & 0.253 & \textbf{3.810}$^\dagger$ & \textbf{3.75} \\
\midrule
\multirow{3}{*}{TLIO dataset}
  & TLIO   & 0.378 & 0.116 & 2.499 & 3.12 \\
  & EqNIO  & \textbf{0.348} & \textbf{0.110} & \textbf{1.850} & \textbf{1.84} \\
  & HeadIO & 0.401 & 0.123 & 2.317 & 2.79 \\
\bottomrule
\end{tabular}
\end{table}

\begin{figure}[t]
    \centering
    \includegraphics[width=\columnwidth]{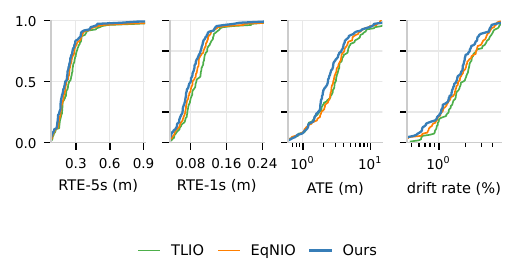}
    \caption{Per-sequence CDFs on Nymeria under ground-truth orientation ($n{=}110$).}
    \vspace{-2mm}
    \label{fig:cdf}
\end{figure}

Table~\ref{tab:main_results} presents our main results on the four datasets. HeadIO achieves the best RTE-5s on all three Aria-glasses datasets, and on the largest benchmark, Nymeria, it achieves state-of-the-art results on all four metrics. Under ground-truth orientation, HeadIO improves RTE-5s by 8.1\% over EqNIO (0.260\,m $\to$ 0.239\,m) and ATE by 8.9\% (3.577\,m $\to$ 3.258\,m), and under IMU-only orientation it maintains a 7.4\% margin (0.297\,m $\to$ 0.275\,m; Table~\ref{tab:ori_robustness}), with the gains consistent across the entire test set (Fig.~\ref{fig:cdf}). On AEA, models trained on Nymeria transfer without fine-tuning. On LaMAria, whose city sequences run up to 30 minutes, HeadIO achieves the best RTE-5s, ATE, and drift, beating TLIO on RTE-5s on every sequence.

On AEA, HeadIO leads RTE-5s by 3.2\% (0.220\,m $\to$ 0.213\,m) and RTE-1s by 4.9\% (0.081\,m $\to$ 0.077\,m), while EqNIO is ahead on ATE by 2.7\% (0.819\,m $\to$ 0.797\,m) and drift rate by 11.5\% (5.81\% $\to$ 5.14\%). The ATE gap comes from a small set of sequences with larger horizontal end-point drift; on the median sequence, HeadIO's ATE is lower by 3.3\% (0.569\,m $\to$ 0.550\,m). On the TLIO dataset, EqNIO leads all metrics; since TLIO does not define a CPF frame, we estimate the forward axis from gravity and walking statistics (Sec.~\ref{sec:datasets}), and HeadIO still outperforms TLIO's own ResNet on ATE by 7.3\% (2.499 $\to$ 2.317) and drift by 10.6\% (3.12\% $\to$ 2.79\%).

\subsection{Ablations}
\label{sec:ablations}

\begin{table}[t]
\centering
\caption{Input representation $\times$ backbone at the 3s window on Nymeria under ground-truth and Mahony (IMU-only) orientation ($n{=}110$, mean over sequences). RTE and ATE in meters, drift in percent; best per column within each backbone in \textbf{bold}.}
\label{tab:prep_backbone}
\setlength{\tabcolsep}{3pt}
\resizebox{\columnwidth}{!}{%
\begin{tabular}{llll cccc cccc}
\toprule
\textbf{Backbone} & \textbf{Frame} & \textbf{Prep.} & \textbf{Prior}
& \multicolumn{4}{c}{\textbf{Ground truth}} & \multicolumn{4}{c}{\textbf{Mahony (IMU-only)}} \\
\cmidrule(lr){5-8}\cmidrule(lr){9-12}
& & & & RTE-5s & RTE-1s & ATE & Drift\% & RTE-5s & RTE-1s & ATE & Drift\% \\
\midrule
\multirow{4}{*}{ResNet}
 & \multirow{2}{*}{World} & --   & TLIO  & 0.278 & 0.101 & 4.093 & 2.17 & 0.310 & 0.108 & 4.514 & 2.20 \\
 &                        & VN   & EqNIO & \textbf{0.260} & \textbf{0.093} & 3.577 & 1.96 & 0.297 & \textbf{0.102} & 4.118 & 2.05 \\
 & \multirow{2}{*}{Body}  & --   & AirIO & 0.285 & 0.103 & 4.806 & 2.68 & 0.314 & 0.109 & 5.493 & 2.88 \\
 &                        & Ours & Ours  & 0.260 & 0.095 & \textbf{3.448} & \textbf{1.84} & \textbf{0.294} & 0.103 & \textbf{3.987} & \textbf{1.96} \\
\midrule
\multirow{4}{*}{Mamba}
 & \multirow{2}{*}{World} & --   & TLIO  & 0.277 & 0.096 & 5.159 & 3.00 & 0.309 & 0.104 & 5.570 & 3.03 \\
 &                        & VN   & EqNIO & \textbf{0.236} & 0.086 & 3.258 & 1.89 & \textbf{0.273} & 0.095 & \textbf{3.820} & \textbf{1.96} \\
 & \multirow{2}{*}{Body}  & --   & AirIO & 0.264 & 0.095 & 3.923 & 2.10 & 0.296 & 0.102 & 4.409 & 2.21 \\
 &                        & Ours & Ours  & 0.239 & \textbf{0.086} & \textbf{3.258} & \textbf{1.79} & 0.275 & \textbf{0.094} & 3.914 & 1.96 \\
\bottomrule
\end{tabular}%
}
\end{table}

\begin{figure}[t]
    \centering
    \includegraphics[width=0.80\columnwidth]{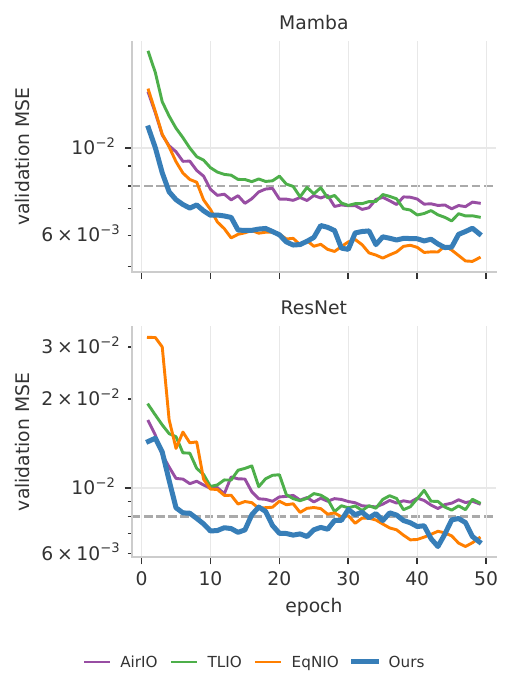}
    \caption{Validation MSE versus training epoch for the four input representations on both backbones (Nymeria, 3s window).}
    \vspace{-2mm}
    \label{fig:convergence}
\end{figure}

In Table~\ref{tab:prep_backbone}, we evaluate the input representation on both backbones at the shared 3s window. Our decomposition improves ATE by 17\% over raw body-frame input on Mamba (3.923 $\to$ 3.258) and by 37\% over the world-frame convention (5.159 $\to$ 3.258), and the same ordering holds on ResNet, where it improves on TLIO's world frame by 16\% (4.093 $\to$ 3.448) and on the full EqNIO pipeline (3.577), indicating that the decomposition generalizes across backbones. Without a suitable representation, the Mamba backbone is not better than a ResNet (world-frame raw ATE 5.159 vs.\ 4.093); with our decomposition, the 0.97M-parameter model leads. Adding EqNIO's VN layers to our Mamba backbone leads RTE-5s by 1.3\% (0.239 $\to$ 0.236) and ties our analytic frame on ATE (3.258 both). Under Mahony orientation it stays ahead (RTE-5s 0.275 $\to$ 0.273), while on ResNet our analytic frame leads RTE-5s by 1.0\% (0.297 $\to$ 0.294), ATE by 3.2\% (4.118 $\to$ 3.987), and drift by 4.4\% (2.05 $\to$ 1.96). The VN layers also cost over 1{,}000\,MFLOPs, more than twice the compute of our entire pipeline (Sec.~\ref{sec:implementation}).

With our decomposition method, the model also converges faster, reaching a common low-error validation level in up to 5$\times$ fewer epochs than the alternative representations on both backbones (Fig.~\ref{fig:convergence}).

\textbf{Window length.} TLIO's default window is 1\,s. Since the Mamba backbone handles longer sequences at linear cost, we varied the window length while keeping the 1s displacement target. A 3s window improves RTE-5s by 21\% over 1\,s (0.304 $\to$ 0.239), while 5\,s brings no further gain (0.244) at 1.7$\times$ the compute cost. We observed similar gains when retraining the ResNet baselines at 3\,s, so we adopt the 3s window for all methods for fair comparison.

\textbf{Backbone ablation.} To verify the choice of Mamba, we also trained a bidirectional LSTM in its place, keeping our heading decomposition and the rest of the network unchanged; despite slightly higher cost (1.06M vs.\ 0.97M parameters, 477 vs.\ 444\,MFLOPs), we found it degrades RTE-5s by 8.4\% (0.239 $\to$ 0.259) under ground-truth orientation.

\begin{figure*}[t]
    \centering
    \includegraphics[width=0.90\linewidth]{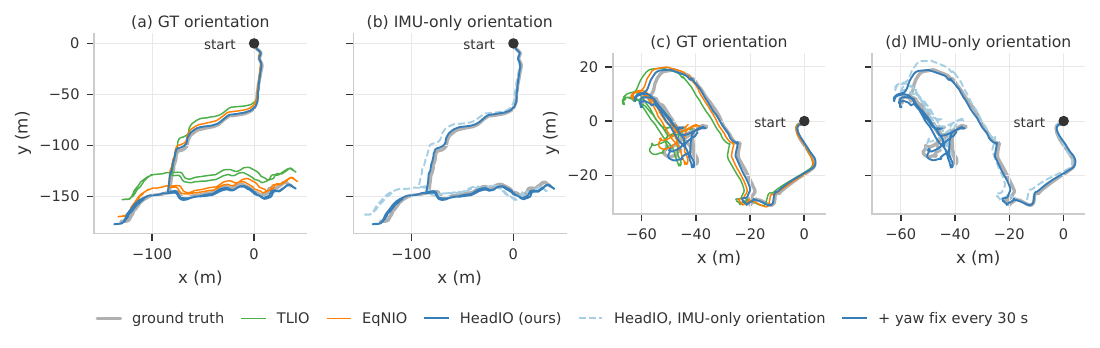}
    \caption{Two Nymeria test sequences. (a,\,c)~Estimated trajectories under ground-truth orientation. (b,\,d)~The same sequences under IMU-only orientation: HeadIO without correction (dashed) and with one external yaw fix every 30\,s (solid).}
    \vspace{-2mm}
    \label{fig:traj_row4}
\end{figure*}

\section{Deployment under Estimated Orientation}
\label{sec:deployment}

\subsection{Robustness to the Orientation Source}
\label{sec:robustness}

\begin{table}[t]
\centering
\caption{Orientation-source study on Nymeria ($n{=}110$, mean over sequences). Each IMU-only source is listed with its own tilt\,/\,yaw error (\textdegree). RTE and ATE in meters, drift in percent; best per metric within each source in \textbf{bold}.}
\label{tab:ori_robustness}
\renewcommand{\arraystretch}{0.90}
\resizebox{\columnwidth}{!}{%
\begin{tabular}{ll cccc}
\toprule
\textbf{Orientation} & \textbf{Method}
& \textbf{RTE-5s} $\downarrow$
& \textbf{RTE-1s} $\downarrow$
& \textbf{ATE} $\downarrow$
& \textbf{Drift\%} $\downarrow$ \\
\midrule
\multirow{3}{*}{Ground truth}
  & TLIO   & 0.278 & 0.101 & 4.093 & 2.17 \\
  & EqNIO  & 0.260 & 0.093 & 3.577 & 1.96 \\
  & HeadIO & \textbf{0.239} & \textbf{0.086} & \textbf{3.258} & \textbf{1.79} \\
\midrule
\multirow{3}{*}{\shortstack[l]{Mahony\\(0.35\,/\,4.21)}}
  & TLIO   & 0.310 & 0.108 & 4.514 & 2.20 \\
  & EqNIO  & 0.297 & 0.102 & 4.118 & 2.05 \\
  & HeadIO & \textbf{0.275} & \textbf{0.094} & \textbf{3.914} & \textbf{1.96} \\
\midrule
\multirow{3}{*}{\shortstack[l]{Madgwick\\(0.93\,/\,4.10)}}
  & TLIO   & 0.335 & 0.120 & 4.575 & 2.18 \\
  & EqNIO  & 0.323 & 0.115 & \textbf{3.963} & \textbf{1.91} \\
  & HeadIO & \textbf{0.299} & \textbf{0.106} & 4.300 & 2.21 \\
\midrule
\multirow{3}{*}{\shortstack[l]{VQF\\(0.99\,/\,4.10)}}
  & TLIO   & 0.333 & 0.118 & 4.707 & 2.31 \\
  & EqNIO  & 0.324 & 0.115 & 4.811 & 2.63 \\
  & HeadIO & \textbf{0.296} & \textbf{0.104} & \textbf{4.435} & \textbf{2.27} \\
\midrule
\multirow{3}{*}{\shortstack[l]{Gyro integration\\(5.52\,/\,4.17)}}
  & TLIO   & 0.697 & 0.182 & 28.535 & 46.30 \\
  & EqNIO  & 0.993 & 0.234 & 57.389 & 82.54 \\
  & HeadIO & \textbf{0.436} & \textbf{0.131} & \textbf{12.645} & \textbf{14.68} \\
\bottomrule
\end{tabular}%
}
\end{table}

In Table~\ref{tab:ori_robustness}, we replace the ground-truth orientation with four IMU-only estimators at evaluation time. All estimators run at 200\,Hz on the same body-frame IMU streams and are initialized from the ground-truth initial orientation; we use the \texttt{ahrs} implementations of Mahony ($k_P{=}0.05$, $k_I{=}0.001$)~\cite{mahony2008nonlinear} and Madgwick ($\beta{=}0.05$)~\cite{madgwick2011estimation}, the \texttt{vqf} implementation of VQF~\cite{laidig2023vqf}, and pure gyro integration. The four estimators have nearly identical yaw error (4.1--4.2\textdegree) but tilt errors spanning over 15$\times$ (0.35\textdegree{} to 5.52\textdegree). We found that degradation tracks the estimator's tilt error rather than its total error. As tilt error grows from 0.35\textdegree{} (Mahony) to 0.93--0.99\textdegree{} (Madgwick and VQF) and to 5.52\textdegree{} (gyro integration), HeadIO's RTE-5s rises from 0.275 to 0.296--0.299 and to 0.436. HeadIO has the best RTE-5s under every source and degrades least under raw gyro integration, with RTE-5s 37\% lower than TLIO (0.697 $\to$ 0.436) and 56\% lower than EqNIO (0.993 $\to$ 0.436). EqNIO degrades most, with ATE growing 16$\times$ from ground truth to gyro (3.577 $\to$ 57.389). Our features are yaw-invariant, as yaw never enters the input (Sec.~\ref{sec:decomposition}); tilt is the only orientation channel that can corrupt the features, and yaw enters only when the predicted displacements are integrated in the world frame.

\subsection{Yaw Correction and Sparse Fixes}
\label{sec:watchdog}

\textbf{Single-coefficient corrector.} Gyroscopes have a small scale-factor error: they report each rotation slightly larger or smaller than it is~\cite{titterton2004strapdown}, so the filter's yaw error grows with the accumulated turn angle, and a head-mounted device turns constantly. On Nymeria the Mahony filter over-counts every turn by 0.49\%, with the same sign on 109 of 110 test sequences; fits on AEA and the TLIO dataset give the same sign and similar magnitude. Subtracting this component with one coefficient, fit on the training split and applied unchanged at test time, reduces mean yaw error by 11.6\% (4.21\textdegree{} $\to$ 3.72\textdegree) and improves RTE-5s by 2.5\% (0.275 $\to$ 0.268). We also retrained a learned gyro calibration network~\cite{brossard2020denoising} on our data; we found it does not improve on this single coefficient.

\textbf{Sparse external fixes.} The remaining yaw error is a random walk and can only be removed by an external reference, but the reference can be sparse. One yaw fix every 30\,s holds the mean yaw error at 0.41\textdegree{} and improves RTE-5s by 12.0\% (0.275 $\to$ 0.242), matching the accuracy with perfect yaw (0.242); a fix every 100\,s still improves it by 10.5\% (0.275 $\to$ 0.246). Figure~\ref{fig:traj_row4}(b,\,d) shows the effect. Without correction, the accumulated yaw error rotates the remainder of the trajectory away from ground truth; with one fix every 30\,s, the estimate can stay aligned. A reference this sparse can come from a magnetometer burst~\cite{li2026mario}, a visual compass~\cite{kosecka2002video}, or occasional relocalization against a known map.

\section{Conclusion}
We presented HeadIO, an analytic body-frame heading decomposition paired with a compact bidirectional Mamba backbone for inertial odometry on head-mounted devices. The decomposition achieves tilt invariance and heading equivariance without learned equivariant layers, composes with any backbone, and gives the 0.97M-parameter model the best RTE-5s in four of five evaluation conditions across four datasets; the exception is the TLIO dataset, which does not define native body-frame semantics. We also found that under IMU-only orientation, accuracy depends mainly on tilt, which the gravity measurement can bound. The systematic part of the yaw error traces to a gyroscope scale-factor term, which a single fitted coefficient removes, and sparse external yaw fixes improve RTE-5s by 12.0\% when given every 30\,s. In the future, non-rigid mounting such as earbuds and intermittent visual-inertial anchoring could extend HeadIO.

\vspace{-3mm}
\bibliographystyle{IEEEtran}
\bibliography{main}

\end{document}